\documentclass{article}

\usepackage[nonatbib,final]{neurips_2021}
\usepackage[numbers]{natbib}

\usepackage[utf8]{inputenc} 
\usepackage[T1]{fontenc}    
\usepackage{hyperref}       
\usepackage{url}            
\usepackage{booktabs}       
\usepackage{amsfonts}       
\usepackage{amsmath}       
\usepackage{amsthm}
\usepackage{nicefrac}       
\usepackage{microtype}      
\usepackage{graphicx}
\usepackage[font=small]{caption}

\usepackage{babel}
\usepackage{textcomp}
\usepackage{subfigure}
\usepackage{bm}
\usepackage{tabularx} 
\usepackage{wrapfig}
\usepackage[ruled,vlined]{algorithm2e}
\usepackage{multicol}
\usepackage{rotating}
\usepackage[export]{adjustbox}
\usepackage{caption}
\usepackage{xurl}
\usepackage[dvipsnames]{xcolor}
\usepackage{wrapfig}

\definecolor{specialgreen}{RGB}{0, 194, 63}

\hypersetup{
    colorlinks=true,
    linkcolor=magenta,
    citecolor=blue,
    filecolor=magenta,      
    urlcolor=blue,
    pdfpagemode=FullScreen,
    }

\newcommand{\PPLR}{PLR$^\parallel$}
\newcommand{\RPLR}{PLR$^\perp$}

\newcommand{\PACCEL}{ACCEL$^\parallel$}

\newcommand{\libname}{\texttt{minimax}}

\newcounter{commentCounter}
\newif\iftrvar
\trvartrue
\iftrvar
\newcommand{\minqi}[1]{{\small \color{blue} \refstepcounter{commentCounter}\textsf{[MJ]$_{\arabic{commentCounter}}$:{#1}}}}
\newcommand{\michael}[1]{{\small \color{cyan} \refstepcounter{commentCounter}\textsf{[MDD]$_{\arabic{commentCounter}}$:{#1}}}}
\newcommand{\jakob}[1]{{\small \color{green} \refstepcounter{commentCounter}\textsf{[JF]$_{\arabic{commentCounter}}$:{#1}}}}

\else
\newcommand{\minqi}[1]{}
\newcommand{\michael}[1]{}
\newcommand{\jakob}[1]{}
\fi

\title{\protect\includegraphics[height=1em]{assets/racing-car-emoji.pdf} \libname{}:\\Efficient Baselines for Autocurricula in JAX}
\title{\libname{}:\\Efficient Baselines for Autocurricula in JAX}

\author{%
  Minqi Jiang \\
  FAIR at Meta AI \\
  \texttt{msj@meta.com} \\
  \And
  Michael Dennis \\
  UC Berkeley \\
  \And
  Edward Grefenstette \\
  UCL \\
  \And
  Tim Rocktäschel \\
  UCL \\
}

\begin{document}

\maketitle

\begin{abstract}
Unsupervised environment design (UED) is a form of automatic curriculum learning for training robust decision-making agents to zero-shot transfer into unseen environments. Such autocurricula have received much interest from the RL community. However, UED experiments, based on CPU rollouts and GPU model updates, have often required several weeks of training. This compute requirement is a major obstacle to rapid innovation for the field. This work introduces the \libname{} library for UED training on accelerated hardware. Using JAX to implement fully-tensorized environments and autocurriculum algorithms, \libname{} allows the entire training loop to be compiled for hardware acceleration. To provide a petri dish for rapid experimentation, \libname{} includes a tensorized grid-world based on \texttt{MiniGrid}, in addition to reusable abstractions for conducting autocurricula in procedurally-generated environments. With these components, \libname{} provides strong UED baselines, including new parallelized variants, which achieve over 120$\times$ speedups in wall time compared to previous implementations when training with equal batch sizes. The \libname{} library is available under the Apache 2.0 license at \url{https://github.com/facebookresearch/minimax}.
\end{abstract}

\section{Introduction and Motivating Work}

\begin{figure}[h]
\centering
\includegraphics[width=0.9\textwidth]{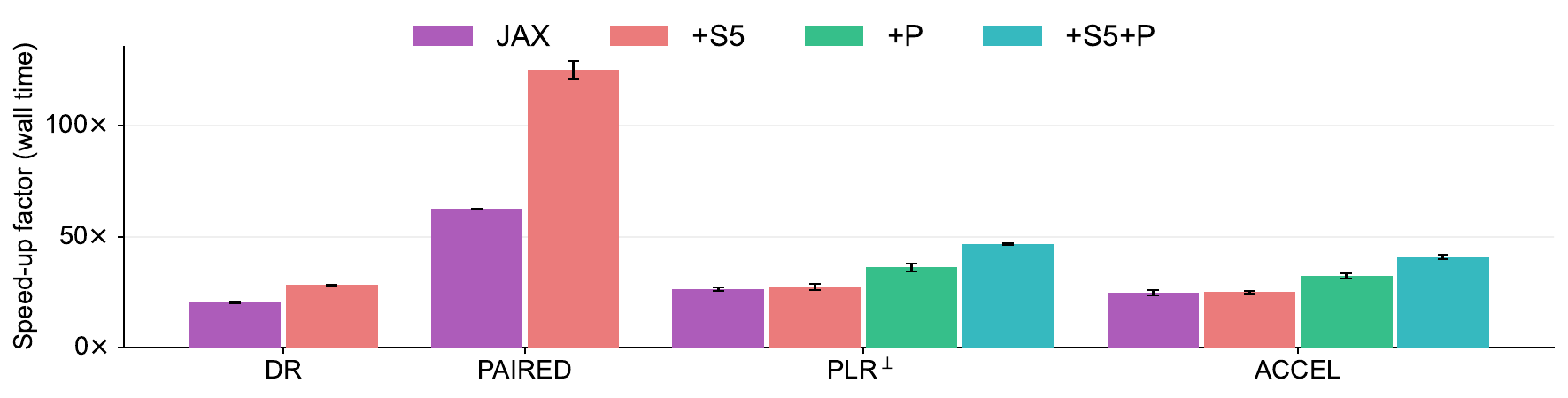}
\caption{Wall time \emph{speed-up factors} achieved by \texttt{\libname{}} relative to PyTorch reference implementations with equal batch sizes (mean and std of 10 runs). Experiments that took 100+ hours now finish in < 3 hours on a single GPU. Here, +S5 indicates the use of an S5 policy, and +P, parallel DCD. (See Section~\ref{sec:baselines} for details.)} 
\label{fig:speedups_summary}
\end{figure}

Autocurricula have proven highly effective in producing powerful deep reinforcement learning (RL) agents in complex multi-agent settings~\citep{silver2017mastering,vinyals2019grandmaster,baker2019emergent,hu2021off,team2021open,samvelyan2023maestro}. Here, a self-organizing curriculum across co-players emerges as agents compete with one another over billions of iterations~\citep{leibo2019autocurricula}. The burgeoning field of \emph{Unsupervised Environment Design}~\citep[UED,][]{dennis2020emergent} recently extends these ideas to the design of the training task itself. In UED, the learning agent or the \emph{student}, plays against a \emph{teacher} in a curriculum game. In each episode of this game, the teacher chooses training tasks or environments in order to maximize some metric based on the student's behavior. A principled choice for the teacher's objective is the student's regret~\citep{savage1951theory,dennis2020emergent}, leading to minimax-regret optimal students should the curriculum game reach an equilibrium.

Many works have demonstrated that UED produces more robust agents in terms of zero-shot transfer to out-of-distribution (OOD) tasks~\citep{dennis2020emergent,jiang2021replay,parker2022evolving,jiang2022grounding, team2023human}. While empirically effective, this class of autocurriculum methods can be computationally expensive. Results on a common maze navigation benchmark, featured in the majority of works on UED, can require on the order of $10^8$ – $10^9$ steps to reproduce. Such experiments can take up to one week to complete, when using the standard UED baselines from the open-source, PyTorch-based \texttt{dcd} codebase~\citep{jiang2021replay} with a V100 GPU---a considerable outlay of compute that is especially taxing for more GPU-poor academic labs. In this technical report, we present \libname{}, a fast and modular library for accelerating the pace of research in UED, as well as other forms of autocurricula~\citep{matiisen2019teacher,wang2019paired,wang2020enhanced,portelas2020teacher, portelas2021automatic,forestier2022intrinsically}. Crucially, \libname{} derives its speed from the powerful JAX library~\citep{frey2023jax} for vector transforms on top of XLA~\citep{leary2017xla}. In addition to a modular library of components for assembling, extending, and evaluating new autocurriculum methods, \libname{} features fully-tensorized implementations of the maze benchmark and standard UED baselines, as well as new parallelized and multi-device variants, achieving over $120\times$ speed-ups in wall time compared to those from \texttt{dcd}, as reported in Figure~\ref{fig:speedups_summary}. 

The main contributions of this work are presented in Sections~\ref{sec:minimax_overview} – \ref{sec:baselines}:
\begin{itemize}
    \item A detailed discussion of the design philosophy and high-level structure of the \libname{} library is provided in Section~\ref{sec:minimax_overview}.

    \item \texttt{AMaze}, a fully-tensorized procedurally-generated, partially-observable maze navigation benchmark for use as a rapid petri dish for evaluating autocurricula is presented in Section~\ref{sec:petri_dish}.

    \item All \libname{} baselines, including novel parallelized and multi-device variants of \RPLR{} and ACCEL, are presented and benchmarked in Section~\ref{sec:baselines}.

\end{itemize}

\section{The \libname{} Library}
\label{sec:minimax_overview}

\begin{figure}[t!]
\centering
\includegraphics[width=1.0\textwidth]{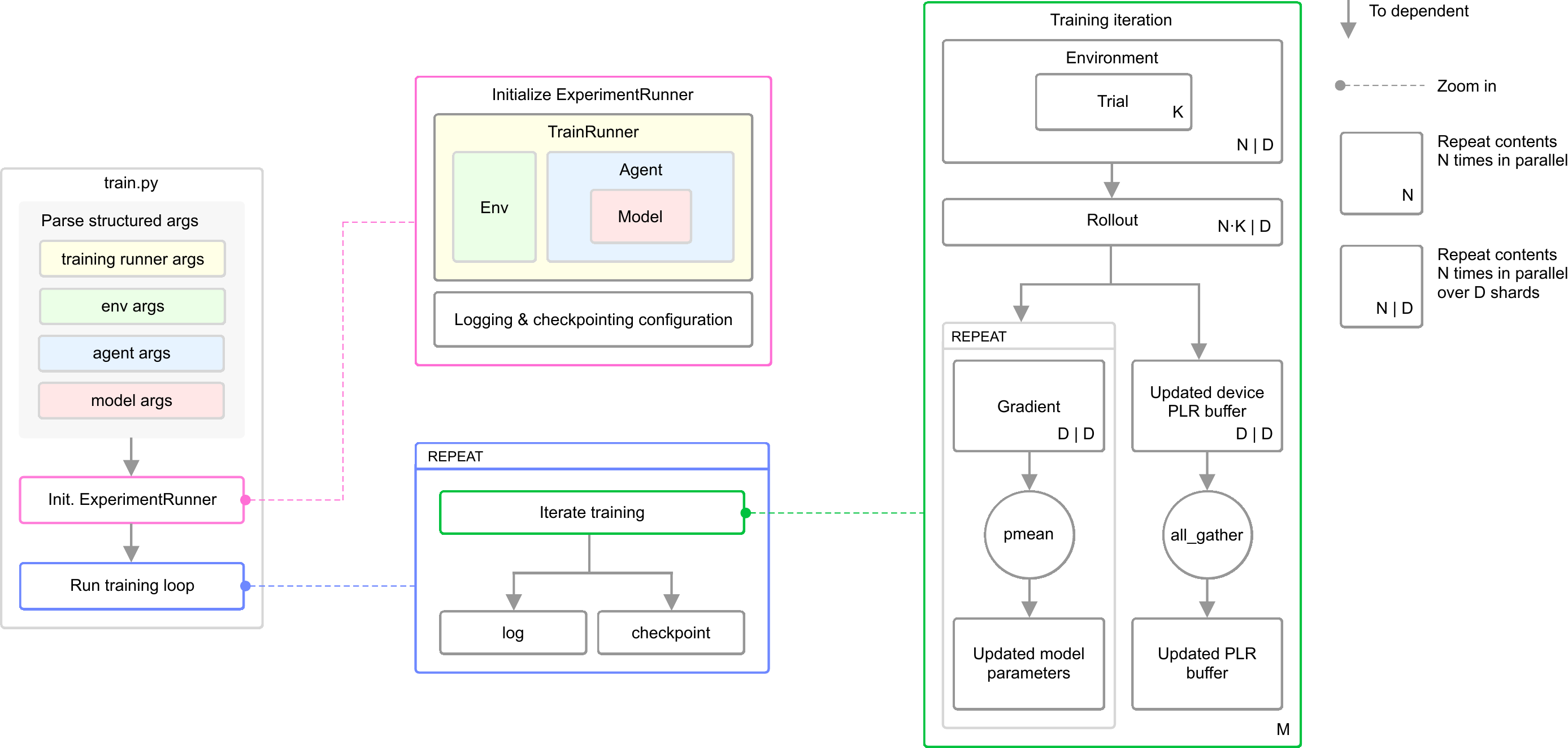}
\caption{A high-level overview of the \texttt{minimax} library. The {\color{specialgreen}training iteration} logic is fully-jitted.}
\label{fig:minimax_system_diagram}
\vspace{-4mm}
\end{figure}

A foundational principle behind \texttt{minimax} is that modularity is crucial for rapid experimentation. Thus, from an architectural standpoint, \libname{} centers around strongly decoupled and broadly interoperable atomic building blocks, consisting of training runners, environments, agents, and models. Each of these building block categories exists as its own submodule. Any component can be easily accessed via a registry interface (similar to the environment registry in \texttt{OpenAI Gym}~\citep{brockman2016openai}), which allows necessary dependencies to be explicitly stated and retrieved, e.g. setting and getting a policy model for a specific environment. In general, dependent components are made generic via dependency injection~\citep{seemann2010dependency} where possible. An overview of \libname{} is presented in Figure~\ref{fig:minimax_system_diagram}.

In order to model the diverse world of autocurriculum methods in a modular way, \libname{} adopts the conceptual structure of \emph{Dual Curriculum Design}~\citep[DCD,][]{jiang2021replay}. DCD describes many autocurricula, including common UED baselines, as unfolding as a game between a student and two kinds of teachers: a generator that can adaptively design tasks and a curator that selectively curates those generated. For example, domain randomization~\citep[DR,][]{tobin2017domain,cad2rl}
corresponds to a random generator (i.e. no adaptation) with no curator; PAIRED~\citep{dennis2020emergent}, to an adaptive generator with no curator; and PLR~\citep{jiang2021prioritized,jiang2021replay} to a curator with a random generator. Each high-level combination of DCD teachers can describe many subsets of UED methods. Thus, \libname{} structures autocurricula as curriculum games and supports extensions to multi-student or multi-teacher games. In the sections that follow, we discuss the role of each of \libname{}'s building blocks in constructing such autocurricula.

\subsection{Runners}
\label{subsec:runners}
Each runner orchestrates a specific kind of curriculum game by coordinating agent and environment components through the registry system. Many autocurricula methods can already be implemented as simple extensions of the runners in the initial release of \libname{}: \texttt{DRRunner}, \texttt{PAIREDRunner}, and \texttt{PLRRunner}, corresponding to DCD autocurricula based on a random teacher, learned generator, and curator respectively. We refer to each algorithmic runner as a \emph{training runner}. A separate \texttt{EvalRunner} performs evaluation of model checkpoints on a prespecified set of test environments.

To minimize configuration overhead, \libname{} follows a highly-declarative approach: A small bundle of logic in a single \texttt{train.py} file executes all algorithms. This design is enabled by a custom argument parser that defines how command-line arguments are packaged into initialization arguments for training runners, agents, models, and evaluators. In practice, this approach pays further dividends in simplifying the addition of new components by removing low-level argument passing from mind. Moreover, the final argument schema serves as a legible blueprint of the corresponding experiment.

Parsed argument groups are passed to their destinations by \texttt{ExperimentRunner}. In addition to managing logging, checkpointing, and periodic calls to \texttt{EvalRunner}, this higher-level runner progresses training by stepping the \texttt{run} method of the training runner, which executes one iteration of the autocurriculum (i.e. one rollout and update cycle for each participating agent). Each \texttt{run} method is fully jittable and can thus be \texttt{vmap} transformed to conduct parallel, independent training runs.\footnote{However this should be avoided for some algorithms to allow \texttt{lax.cond} to perform efficient branching, e.g. as used in \RPLR{} updates, that is currently foregone by XLA once inside \texttt{vmap}.}

\subsection{Environments}
\label{subsec:envs}

The \libname{} library environment interface is based on that from the \texttt{Gymnax}~\citep{lange2022gymnax} framework, but departs in several ways: First, \libname{} supports environment wrappers, which carry their own state. This state, is passed across timesteps via an additional \texttt{extra} dictionary in the tuple returned by environments' \texttt{step} and \texttt{reset} methods. This design shields wrapper-specific metadata from environment-specific metadata, typically conveyed within the \texttt{info} component of the tuple. Second, to measure how distributions of environment metrics evolve over time, \libname{} environments can optionally implement the \texttt{get\_env\_metrics} method, which returns a dictionary of environment attributes, e.g. number of walls in a maze. We now describe two more consequential design choices in how \libname{} handles environments. 

\paragraph{Hierarchical parallelism} Usefully, \libname{} directly supports environment parallelism across a hierarchy via the \texttt{BatchEnv} decorator class: agents (i.e. the population batch), evaluations, and environments. Specifically, the last two environment batch dimensions are flattened inside \texttt{BatchEnv} instances. This grouping allows tidy implementations of multi-student or multi-teacher training logic, as well as parallelizing environment logic across not only multiple environment instances, but also evaluations of each instance, e.g. to obtain a denoised estimate per instance. Crucially, environment logic is specified, as normally, for a single instance, and all parallelism is abstracted into \texttt{BatchEnv}.

\paragraph{UPOMDP as a first-class citizen} Prior RL libraries are designed to support standard (partially-observable) Markov decision processes~\citep[MDPs,][]{puterman1994markov}. However, autocurriculum methods typically operate over an extension, called an \emph{Underspecified POMDP}~\citep[UPOMDP,][]{dennis2020emergent}, which explicitly considers the set of configurable free parameters of the environment, which these prior libraries ignore. In \libname{}, environment parameters are separated into \emph{static parameters}, which define fixed aspects set at initialization, e.g. maze size, and free parameters, which vary per instance and are stored in the environment state. Thus, to fully exploit free parameters in generating autocurricula, \libname{} environments implement getter and setter methods for the environment state.

UED with a learned adversary, e.g. PAIRED, requires the teacher to make a sequence of design decisions in a separate MDP that results in producing a specific environment instance. To directly model such teacher MDPs, \libname{} implements the teacher's decision process as its own \texttt{Environment} subclass, which is directly registered to match with the student's corresponding \texttt{Environment} class. The \texttt{BatchUEDEnv} decorator class then takes the pair of environments and vectorizes a shared student and teacher interface called \texttt{UEDEnvironment}: Once the teacher steps through its rollout, calling this interface's \texttt{reset\_student} method sets the student's UPOMDP to the final instance designed by the teacher. This approach allows for cleaner, modular development of student and teacher decision processes, while allowing for a simple, shared environment abstraction for use in training runners.

\vspace{-1mm}
\subsection{Agents and Models}
\label{subsec:agents}

In \libname{}, an agent corresponds to a specific algorithm for optimizing a sequential-decision making policy, while a model corresponds to a module implementing the underlying policy. Following a common pattern in \libname{}, models are dependency-injected into \texttt{Agent} instances. This design allows the same implementation of a \texttt{PPOAgent} to perform PPO over any policy model for any environment. In general, training runners operate over populations of agents, and any individual \texttt{Agent} can be transformed into a population via the \texttt{AgentPop} decorator class. Combined with the batch environment classes, these abstractions make it simple to extend future releases of \texttt{minimax} to support more complex multi-agent settings~\citep{pbt,lanctot2017unified,vinyals2019grandmaster,samvelyan2022maestro}.

\section{Maze as an Accelerated Petri Dish}
\label{sec:petri_dish}

Mazes are a simple and effective setting for studying autocurricula: Mazes exhibit interpretable difficulty gradients (in terms of initial distance to the goal and number of obstacles) and agent behavior. Moreover, the combinatorial design space of potential obstacle configurations serves as a rich and easily extensible domain for UED. Importantly, many difficult problems in RL, such as decision-making under partial observability and generalization to unseen task instances can be directly studied in this domain. A large share of prior works on autocurricula have thus used 2D mazes to initially vet algorithmic ideas and other hypotheses~\citep{igl2019generalization, raileanu2020ride, amigo, zhang2020bebold, dennis2020emergent,jiang2021prioritized}. 

\begin{wraptable}{r}{0.5\textwidth}
\vspace{-4mm}
\small
\begin{center}
\caption{Speed comparison of \texttt{AMaze} and \texttt{MiniGrid} mazes for varying degrees of parallelism.}
\label{table:maze_speed}
\scalebox{0.9}{
\begin{tabular}{ llllr } 
\toprule
\# Environments & \textbf{1} & \textbf{32} & \textbf{256} & \textbf{1024}   \\ 
\midrule
 \emph{Steps-per-second (SPS)} \\
\texttt{MiniGrid} & 2k & 12k & 16k & 16k \\
\texttt{AMaze} & 3k & 76k & 582k & 2M  \\
\midrule
Speedup & 1.6$\times$ & 6.5$\times$ & 37$\times$ & 136$\times$ \\
\bottomrule
\end{tabular}}
\end{center}
\vspace{-4mm}
\end{wraptable}

\begin{figure}[t!]
\centering
\includegraphics[width=1.0\textwidth]{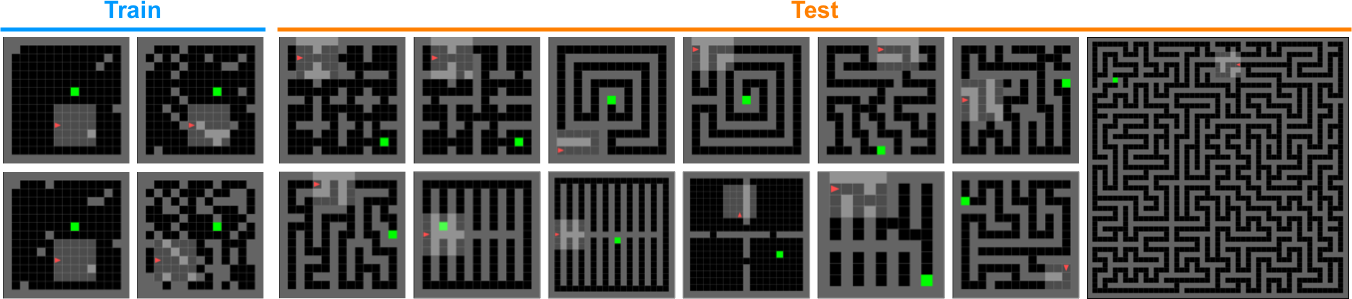}
\caption{Example training and test environments in \texttt{AMaze}, a fully-tensorized maze environment in JAX.}
\label{fig:maze_overview}
\vspace{0mm}
\end{figure}

Thus, we include a procedurally-generated 2D maze environment in \texttt{minimax}. This environment, which we call \texttt{AMaze}, is a fully-tensorized implementation of goal-reaching mazes based on \texttt{MiniGrid}~\citep{minigrid}. In its basic form, \texttt{AMaze} provides a challenging, partially-observable navigation task for RL agents. It can be configured to any rectangular grid layout with randomly sampled wall, goal, and agent positions (see Figure~\ref{fig:maze_overview}). Further, \texttt{AMaze} is fully compatible with the \texttt{UEDEnvironment} abstraction in \libname{}, allowing full support for UED with learned teachers and directly setting free parameters (i.e. the tile map). Most importantly, \texttt{AMaze} is fast. As seen in Table~\ref{table:maze_speed}, even with an environment batch size of 1, \texttt{AMaze} runs $60\%$ faster in terms of steps per second (SPS). As environment parallelism increases, \texttt{AMaze} takes greater advantage of XLA parallelism, achieving speedup factors of $136\times$ at 1024 parallel environments. 

In order to compute shortest paths within a fully-jitted \texttt{run} method, \texttt{AMaze} makes use of Seidel's Algorithm~\citep{seidel1995all}, a matrix-based all pairs shortest path algorithm with time complexity $O(V^{\omega}\log V)$. Here, $V$ is the number of vertices and $\omega < 2.373$ is the exponent in the computational complexity of matrix multiplication. This all-pairs approach further enables efficiently computing more exotic complexity metrics such as shortest-paths distributions~\citep{xland} or resistance distance distributions~\citep{randic1993resistance}.

Importantly, \texttt{AMaze} exactly replicates the \texttt{MiniGrid}-based mazes in prior works, including the full OOD benchmark~\citep{jiang2021replay}, enabling direct comparisons with \libname{} implementations.

\section{Efficient Baselines}
\label{sec:baselines}

The central aim of \libname{} is to provide fast open-source implementations of strong autocurriculum baselines. In this way, we seek to remove the computational overhead for researchers to collectively innovate on this exciting algorithmic frontier. Table~\ref{table:baselines} summarizes the current baselines in \libname{}. Notably, we introduce new parallelized variants of \RPLR{} and ACCEL{} that, as described later in this section, achieve additional wall-time gains compared to the PyTorch~\citep{paszke2019pytorch} reference implementations~\citep{jiang2021replay,parker2022evolving}. All evaluations used V100 GPUs and Intel Xeon E5-2698 v4 CPUs.

\begin{table*}[t!]
\small
\begin{center}
\caption{\small{Comparison of wall time and task performance (mean and std of 10 runs) between \libname{} and \texttt{dcd}, based on training for 30k PPO updates. Corresponding runs use equal PLR replay rate and PPO minibatch and epoch settings. \PPLR{} and \PACCEL{} compare to \RPLR{} and ACCEL in \texttt{dcd} respectively.}}
\label{table:relative_perf}
\scalebox{0.9}{
\begin{tabular}{ rrrrrrrr } 
\toprule
 & \textbf{DR} & \textbf{PAIRED} & \textbf{\RPLR{}} & \textbf{\PPLR{}}  & \textbf{ACCEL} & \textbf{\PACCEL{}} \\
 \midrule
\emph{Wall time (hours)} \\
\texttt{dcd} & $63\pm2$ & $426\pm47$ & $119\pm0$ & – & $104\pm8$ & – \\
\libname{} & $3\pm0$ & $7\pm0$ & $5\pm0$ & $3\pm0$ & $4\pm0$ & $3\pm0$ \\
\midrule
Speedup & $20\times$ & $62\times$ & $26\times$ & $36\times$ & $24\times$ & $32\times$ \\
\midrule
\emph{Solved rate} \\
\texttt{dcd} & $0.62\pm0.05$ & $0.52\pm0.13$ & $0.71\pm0.04$ & – & $0.75\pm0.03$ & – \\
\libname{} & $0.55\pm0.05$ & $0.63\pm0.04$ & $0.82\pm0.02$ & $0.80\times0.02$ & $0.83\pm0.02$ & $0.78\pm0.03$ \\
\midrule
Relative solved rate & $0.88\times$ & $1.22\times$ & $1.16\times$ & $1.12\times$ & $1.12\times$ & $1.06\times$ \\
\bottomrule
\end{tabular}
}
\end{center}
\end{table*}

\begin{wraptable}{r}{0.5\textwidth}
\vspace{-6mm}
\small
\begin{center}
\caption{Baselines currently implemented in \libname{}.}
\label{table:baselines}
\label{table:algos}
\scalebox{0.85}{
\begin{tabular}{ llr } 
\toprule
\textbf{Algorithm}   & \textbf{Reference} & \textbf{Runner}  \\ \midrule
DR  & Tobin et al, 2019~\citep{tobin2017domain} & \texttt{dr}  \\ 
Minimax UED & Dennis et al, 2020~\citep{dennis2020emergent} & \texttt{paired}  \\ 
PAIRED  & Dennis et al, 2020~\citep{dennis2020emergent} & \texttt{paired}  \\ 
Pop. PAIRED  & Dennis et al, 2020~\citep{dennis2020emergent} & \texttt{paired}  \\ 
PLR  & Jiang et al, 2021~\citep{jiang2021prioritized} & \texttt{plr}  \\
Robust PLR  & Jiang et al, 2021~\citep{jiang2021replay} & \texttt{plr}  \\
ACCEL  & Parker-Holder et al, 2022~\citep{parker2022evolving} & \texttt{plr}  \\
Parallel PLR  & Introduced in this work & \texttt{plr}  \\
Parallel ACCEL & Introduced in this work & \texttt{plr}  \\
\bottomrule
\end{tabular}}
\end{center}
\vspace{-4mm}
\end{wraptable}

\begin{wrapfigure}{r}{0.5\textwidth}
\vspace{-5mm}
\centering
\includegraphics[width=0.5\textwidth]{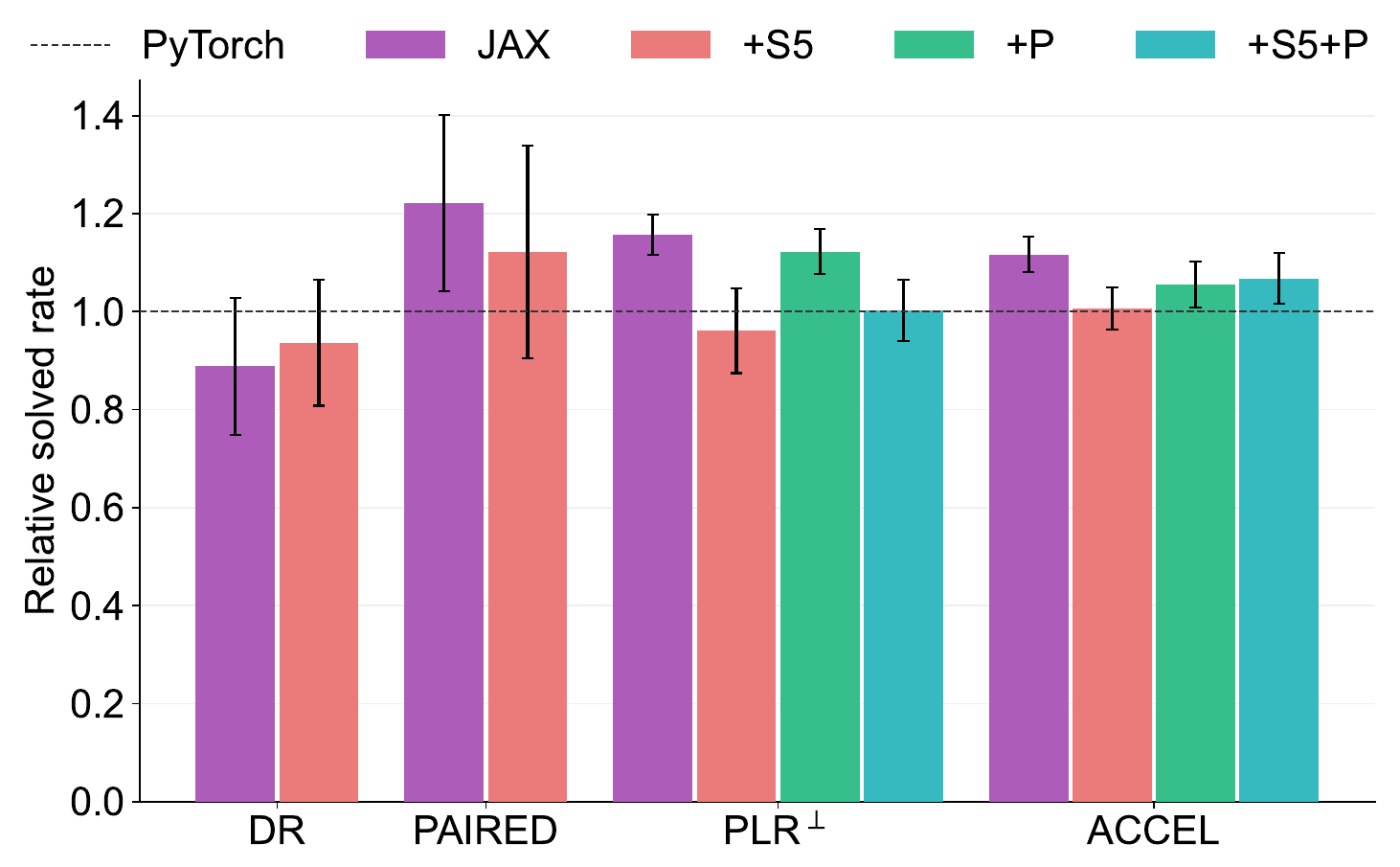}
\caption{\small{Relative solved rates across \texttt{AMaze} test mazes compared to PyTorch \texttt{dcd} (mean and std of 10 runs). The label +P indicates parallel DCD.}}
\label{fig:relative_solved_rate}
\vspace{-4mm}
\end{wrapfigure}

\paragraph{UED baselines}
Currently \libname{} contains baselines that are
variants on three core algorithms: domain randomization (DR), PAIRED, and Prioritized Level Replay (PLR). Important variants of these include Robust PLR (\RPLR{}), which only updates the agent after replay episodes; ACCEL, which replaces \RPLR{}'s random search with evolutionary search over the level buffer; and Population PAIRED, which uses a learned teacher to maximize relative population regret, as defined by the maximum return achieved by any individual minus the mean performance across the population of $N \ge 2$ students. Our PAIRED evaluations compare to the stronger ``high-entropy'' baseline~\citep{mediratta2023stabilizing}. In \libname{}, the main rollout and update loop of each algorithm is fully jitted via JAX, and environment parallelism is accomplished via the vectorizing decorator classes described in Section~\ref{sec:minimax_overview}. Importantly, \libname{} avoids the computational overhead of the multiple student rollouts required by PAIRED variants by parallelizing all student rollouts. Under equal batch sizes, these optimizations result in consistent wall time speedups of at least $20\times$ or higher across all baselines, in comparison to the \texttt{dcd} library, the previous reference implementation in PyTorch (see Tables~\ref{table:relative_perf} and \ref{table:relative_perf_s5}), while maintaining comparable test performance on the full OOD maze benchmark (see Figure~\ref{fig:relative_solved_rate}). Practically speaking, these speedups mean that experiments that once took hundreds of hours to finish can now be done in just a few hours on a single GPU. Figure~\ref{fig:benchmark_abs_solved_rate} in Appendix~\ref{app:experiments} compares the absolute solved rate of each method. Appendix~\ref{app:hps} details the choice of hyperparameters and model architectures.

\begin{wrapfigure}{r}{0.4\textwidth}
\vspace{-1mm}
\centering
\includegraphics[width=0.4\textwidth]{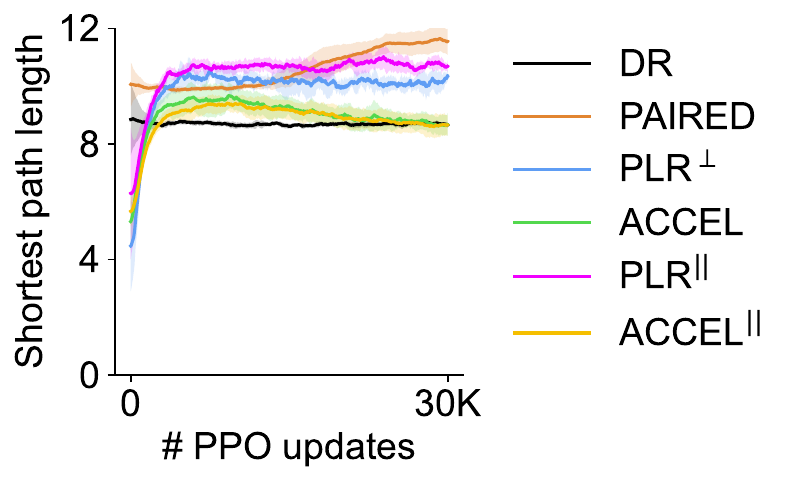}
\caption{\small{Shortest path lengths of training mazes per method (mean and std of 10 runs}).}
\label{fig:complexity_shortest_paths}
\vspace{-8mm}
\end{wrapfigure}

\begin{figure}[t!]
\centering
\includegraphics[width=1.0\textwidth]{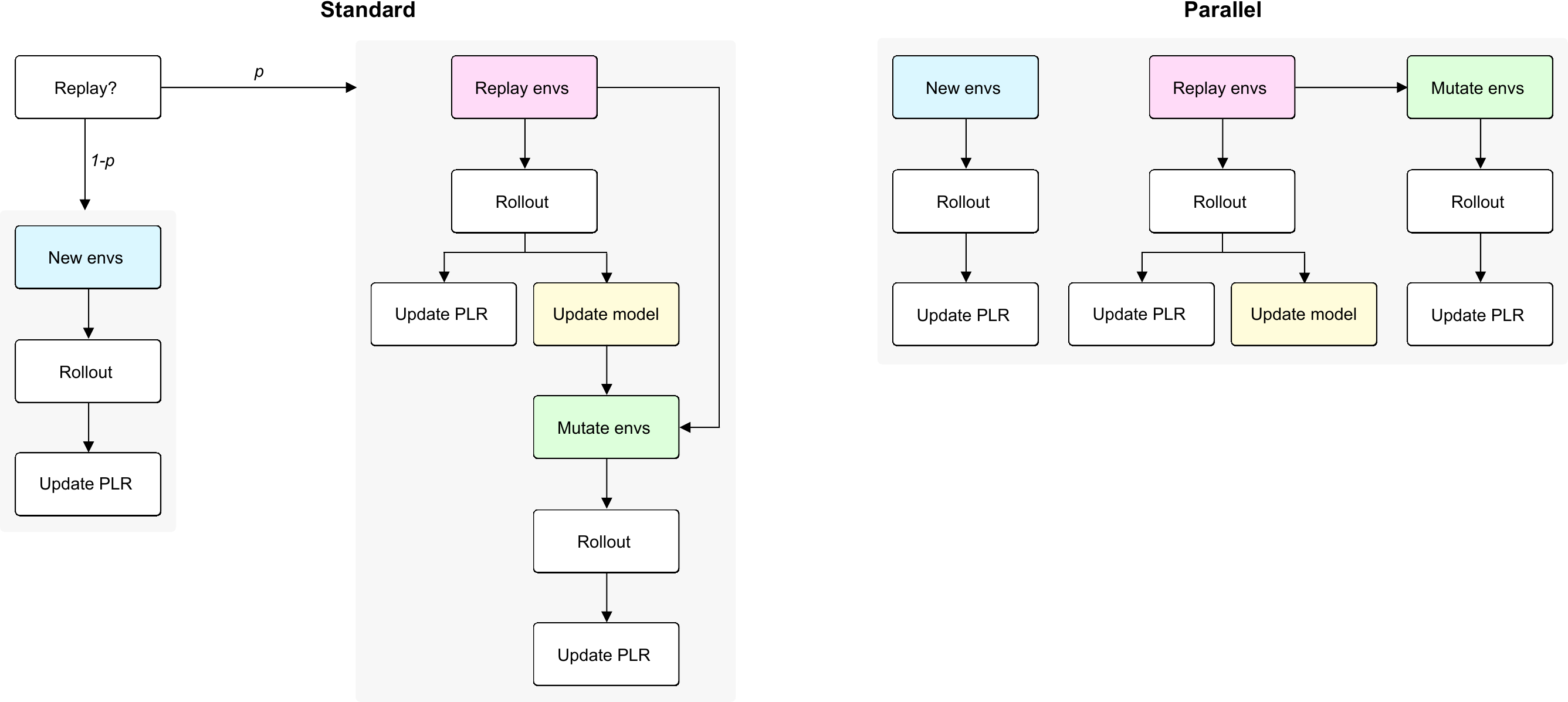}
\caption{Left: The sequence of operations in standard implementations of \RPLR{} and ACCEL. Right: \PPLR{} and \PACCEL{} reduce wall time by executing rollouts for new levels, replay levels, and mutated levels in parallel.}
\label{fig:parallel_dcd_overview}
\end{figure}

\paragraph{Parallel DCD} Next, we seek to push speedups even higher by introducing  Parallel PLR (\PPLR{}) and Parallel ACCEL (\PACCEL{}). The key insight behind these two methods is that both search and replay steps of \RPLR-based methods can be run fully in parallel, allowing these methods to take full advantage of hardware-accelerated parallelism. Specifically, \PPLR{} doubles the environment batch size, filling the first half with newly-sampled environment instances and the second, with replay levels. \PPLR{} then evaluates and updates its level buffer with this full batch. \PACCEL{} follows exactly the same approach with a tripled environment batch size. The first two thirds of the buffer are evenly split between newly sampled and replay levels, while the last third consists of mutations of replay levels from the same batch (see Figure~\ref{fig:parallel_dcd_overview}). We find that \PPLR{} and \PACCEL{} show wall time speedups compared to their standard counterparts of $38\%$ and $33\%$ respectively (see Table~\ref{table:relative_perf}). This boost results in these replay-based UED matching simple DR in wall time. Figure~\ref{fig:complexity_shortest_paths} shows that the parallel variants also ratchet up the shortest path length as training progresses, with \PPLR{} notably overtaking \RPLR{}.

\begin{figure}[t!]
  \centering
  \subfigure{
    \includegraphics[width=0.4\linewidth]{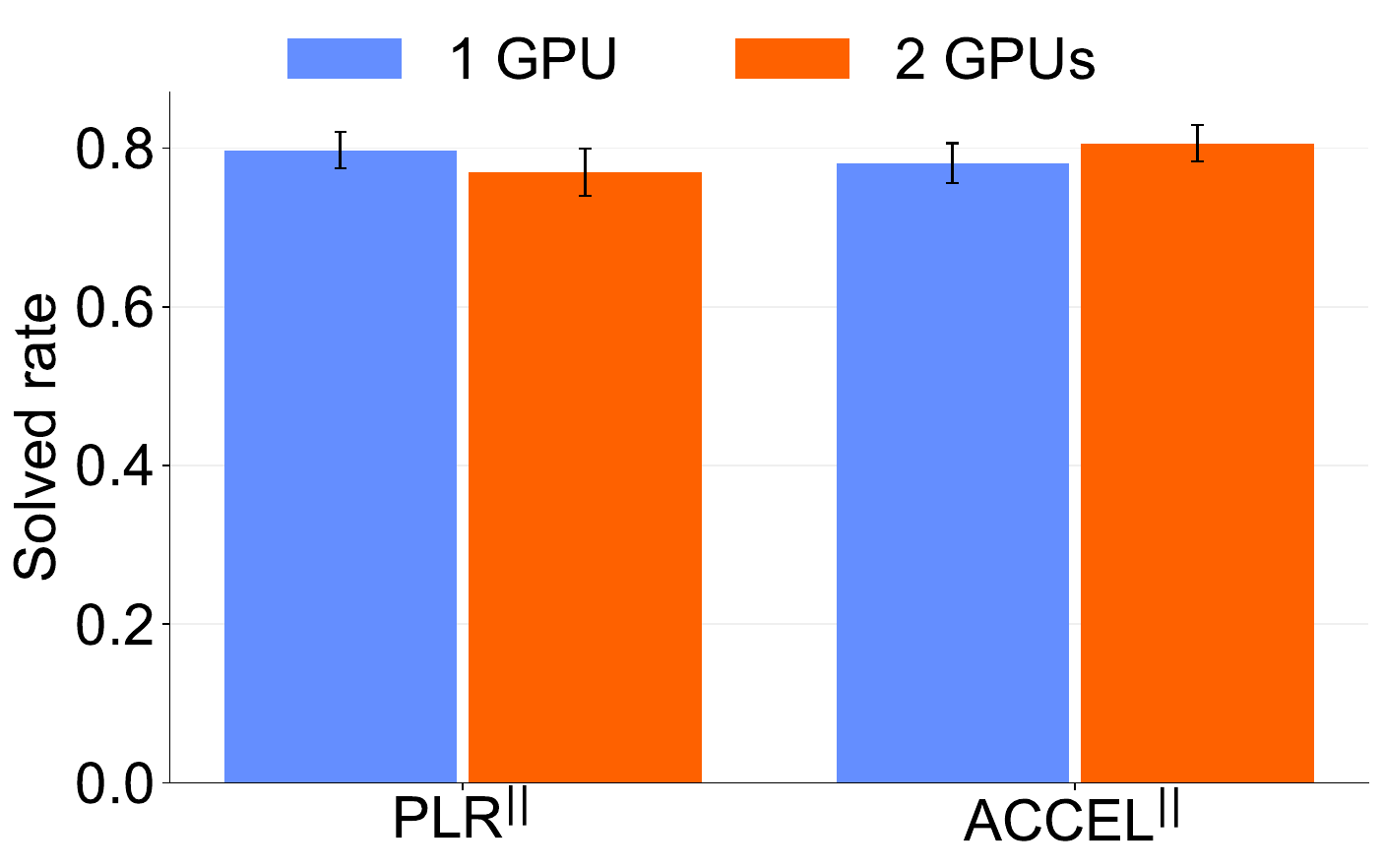}
  }
  \hspace{0.5cm}
  \subfigure{
    \includegraphics[width=0.4\linewidth]{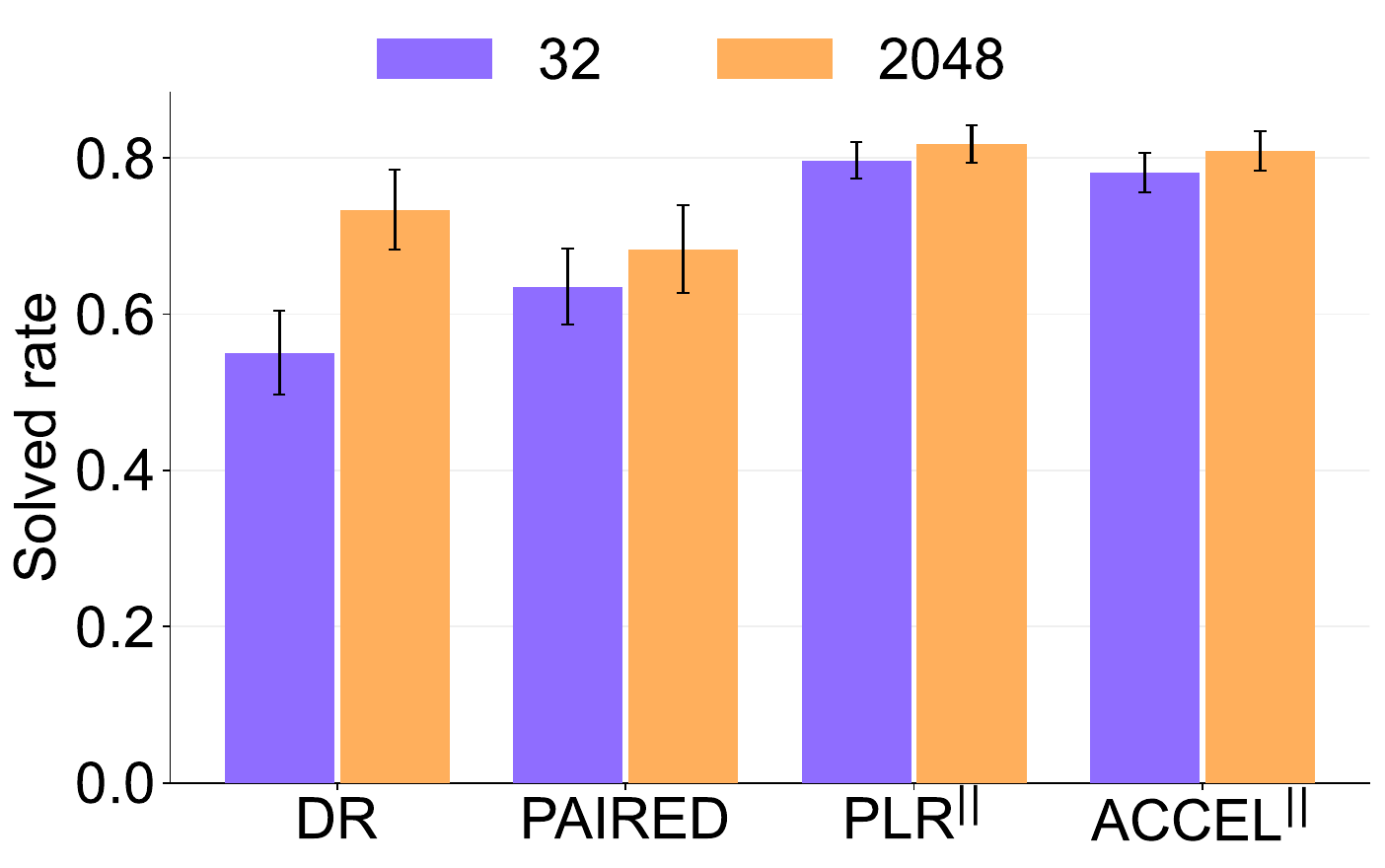}
  }
  \caption{\small{Solved rate over the full OOD maze benchmark for \PPLR{} and \PACCEL{} using 32 parallel environments split across either one or two GPUs (left) and for each method for 32 vs. 2048 parallel environments (right). The plots show means and std across 10 training runs.}}
  \label{fig:solved_rate_vs_parallelism}
\end{figure}

\paragraph{S5 policies} We replace the LSTM~\citep{hochreiter1997long} in the student policy with a stack of S5 layers~\citep{smith2022simplified}. S5 is a variant of the structured state-space sequence model~\citep{gu2021efficiently}, which is $O(\log L)$ in the backward pass for a sequence of length $L$. S5 has been shown to significantly outperform LSTMs in both wall-time efficiency during training and performance on RL tasks requiring conditioning on a long history~\citep{lu2023structured}. We report the wall-time speedups and performance on the full OOD maze benchmark in Figure~\ref{fig:speedups_summary} and Table~\ref{table:relative_perf_s5}. The performance of \libname{} baselines, for both S5 and LSTM policies, relative to the corresponding LSTM-based PyTorch reference implementation in \texttt{dcd}, is shown in  Figure~\ref{fig:relative_solved_rate}. 

While S5 policies match the test performance of LSTM policies, they tend to exhibit greater sensitivity to hyperparameters, like learning rate, and higher variance across training runs. We find that, unlike in Lu et al, 2022~\citep{lu2023structured}, which focuses on continuous-control tasks, applying layer normalization~\citep{ba2016layer} to either the input or output of each S5 layer is essential for matching LSTM test performance in the maze domain. (Maze navigation episodes are limited to 250 time steps, which can explain why S5 fails to outperform LSTM in test performance.) Likely, other design choices in the S5 architecture will lead to further improvements in generalization performance and reduced sensitivity to hyperparameters. Given the wall-time speedups at training when switching to policies based on structured state-space models, principled modifications of this architectural family for the RL setting are a promising direction for future research.

\begin{table*}[t!]
\small
\begin{center}
\caption{\small{Comparison of wall time and task performance (mean and std of 10 runs) between \libname{} and \texttt{dcd} with S5 policies, based on training for 30k PPO updates. Corresponding runs use equal PLR replay rate and PPO minibatch and epoch settings. \PPLR{}+S5 and \PACCEL{}+S5 compare to \RPLR{} and ACCEL in \texttt{dcd} respectively.}}
\label{table:relative_perf_s5}
\scalebox{0.9}{
\begin{tabular}{ rrrrrrrr } 
\toprule
& \textbf{DR} & \textbf{PAIRED} & \textbf{\RPLR{}} & \textbf{\PPLR{}}  & \textbf{ACCEL} & \textbf{\PACCEL{}} \\
 \midrule
\emph{Wall time (hours)} \\
\texttt{dcd} & $63\pm2$ & $426\pm47$ & $119\pm0$ & – & $104\pm8$ & – \\
\libname{} + S5 policy & $2\pm0$ & $3\pm0$ & $4\pm0$ & $3\pm0$ & $4\pm0$ & $3\pm0$ \\
\midrule
Speedup & $28\times$ & $125\times$ & $27\times$ & $47\times$ & $25\times$ & $40\times$ \\
\midrule
\emph{Solved rate} \\
\texttt{dcd} & $0.62\pm0.05$ & $0.52\pm0.13$ & $0.71\pm0.04$ & – & $0.75\pm0.03$ & – \\
\libname{} + S5 policy & $0.58\pm0.05$ & $0.58\pm0.06$ & $0.68\pm0.04$ & $0.71\times0.03$ & $0.74\pm0.02$ & $0.79\pm0.03$ \\
\midrule
Relative solved rate & $0.94\times$ & $1.12\times$ & $0.96\times$ & $1.00\times$ & $1.00\times$ & $1.06\times$ \\
\bottomrule
\end{tabular}
}
\end{center}
\end{table*}

\paragraph{Multi-device training} All training runners in \libname{} support sharding environment rollouts and gradient computation across multiple devices. A high-level diagram of how each step of the training runner's update cycle is sharded along the environment batch dimension across devices is shown in Figure~\ref{fig:minimax_system_diagram}. 
Notably, the standard implementations of PLR-based methods like \RPLR{} and ACCEL entail an \texttt{all\_gather} bottleneck when updating the PLR buffer. We avoid this issue by introducing a new variant of these methods, whereby separate PLR buffers are maintained per device throughout training, effectively sharding the buffer across devices, such that the sum of all individual buffer sizes equals the original, unsharded buffer size. For $D$ devices and an unsharded PLR buffer of size $B$, this approach results in $D$ independent PLR buffers, each of size $B/D$ and updated with only the environment instances run on its own shard. Figure~\ref{fig:solved_rate_vs_parallelism} shows that these \emph{synchronous data-parallel} (SDP) variants and the standard PLR-based methods result in comparable zero-shot transfer performance on the full OOD maze benchmark. 

\begin{figure}[t!]
  \centering
  \includegraphics[width=1.0\textwidth]{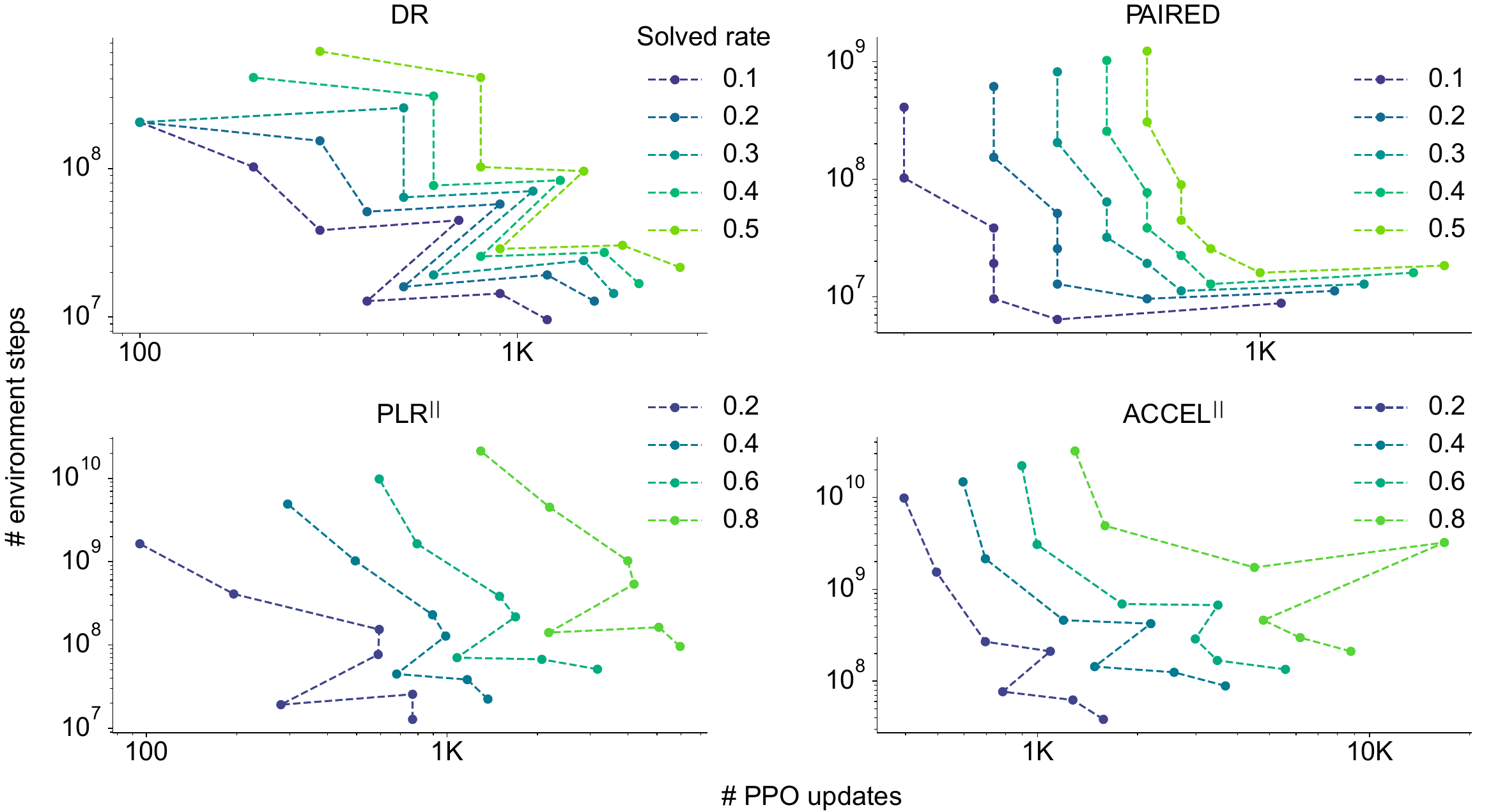}
  \caption{Minimum number of environment transitions and PPO updates needed to achieve a specific mean solved rate on validation mazes. All axes are in log scale. Each point is a mean across 10 training runs.}
  \label{fig:batch_size_pareto_curves}
\end{figure}

\paragraph{Batch-size scaling} The multi-device baselines in \libname{} allow straightforward scaling to much larger environment batch sizes via the \texttt{shmap} (shard map) transform in JAX. We investigate the zero-shot transfer performance of DR, PAIRED, \PPLR{}, and \PACCEL{} on OOD test mazes with increasing training batch sizes. Figure~\ref{fig:solved_rate_vs_parallelism} shows that increasing the environment batch size to 2048 parallel environments leads to significant improvements in zero-shot solved rates over the full OOD maze benchmark. This phenomenon aligns with previous theoretical and empirical observations of how increasing training batch size results in improved signal-to-noise ratio in the stochastic gradient estimates~\citep{mccandlish2018empirical}. This results in a general trend in which larger batch sizes tend to require fewer updates to achieve the same degree of performance, while requiring more total number of samples (in this case, environment transitions). Figure~\ref{fig:batch_size_pareto_curves} visualizes this trade-off by plotting the empirical relationship between the minimum number of environment transitions and minimum number of PPO updates required to reach a fixed degree of performance on three fixed OOD validation mazes, based on training with 32, 64, 128, 256, 512, 1024, and 2048 parallel environments per rollout. Here, each fixed performance curve tends to move from upper left to lower right, reflecting how larger batch sizes tend to be more update efficient, but less sample efficient. These results show that, independent of more sophisticated methodological changes, simply scaling the training batch size can result in significant improvements in OOD test performance when training with autocurriculum methods.

\section{Related Work}
Many recent works implement fast, GPU-accelerated RL algorithms and environments in JAX~\citep{bradbury2021jax}. They include Brax~\citep{freeman2021brax}, a differentiable physics environment; Jumanji~\citep{bonnet2023jumanji}, a collection of combinatorial optimization problems; gymnax~\citep{lange2022gymnax}, a JAX library for single-agent RL training with ports of the popular MinAtar suite~\citep{young2019minatar} and BSuite~\citep{osband2019behaviour}; JaxMARL~\citep{rutherford2023jaxmarl}, a collection of JAX-based multi-agent RL environments; \texttt{EvoJAX}~\citep{tang2022evojax} and \texttt{evosax}~\citep{lange2023evosax} for evolutionary optimization; and PureJAXRL~\citep{lu2023purejaxrl,lu2022discovered}, a minimalist framework for both evolutionary optimization and RL. Several previous works have also considered optimizing the speed of experience collection by implementing fast, asynchronous rollout workers~\citep{stooke2018accelerated, berner2019dota, espeholt2018impala, kuttler2019torchbeast, hessel2021podracer}. This approach has been combined with highly GPU-optimized environment implementations to achieve impressive throughput~\citep{bamford2021griddly, petrenko2021megaverse, shacklett23madrona}.

\section{Discussion}

We introduced \libname{}, a fast JAX-based library that enables rapid experimentation in autocurriculum research for RL. By driving down computational costs, we hope our work can accelerate further progress in this exciting field. We highlighted key features of \libname{}, namely its modular structure and its associated experimental playground for rapid iteration---a fully-tensorized, hardware-accelerated procedural maze environment. Building on these components, our JAX-based autocurriculum baselines, including new parallelized and multi-device variants, resulted in training runs that were up to $120\times$ faster in wall time than the PyTorch reference implementations under the same training batch size. Experiments that once required hundreds of hours of compute can now finish in just a couple of hours on a single GPU. Crucially, our evaluations demonstrated that the \libname{} baselines either match or exceed the test performance of previous reference implementations, while drastically shrinking the timescale of autocurriculum research.

\newpage 

\section*{Author Contributions}
MJ led the project from conception. He designed and implemented the library, formulated the new parallel variants of \RPLR{} and ACCEL, conducted the experiments, and led paper writing. MD, EG, and TR provided invaluable feedback and suggestions. MD also assisted with the design of the example notebooks. EG first broached the idea of parallelizing the new and replay level evaluation branches in PLR. TR strongly encouraged writing this manuscript to accompany the code release.

\ack{We thank Chris Lu and Robert Lange for their work on scaling up JAX-based RL, which greatly influenced \libname{}. We further thank Chris Lu and Ben Ellis for conversations that benefited this work. We also thank Michael Beukman and Alexander Rutherford for finding important bugs in the initial release. Releasing \texttt{minimax} under the Apache 2.0 license would not be possible without the support of Leon Bottou, Naila Murray, and Kerry Andken at Meta AI. MJ is funded by Meta AI.}


\bibliographystyle{abbrv}
\bibliography{refs}

\appendix

\newpage

\section{Additional Experiment Results}
\label{app:experiments}

\begin{figure}[h]
\centering
\includegraphics[width=0.8\textwidth]{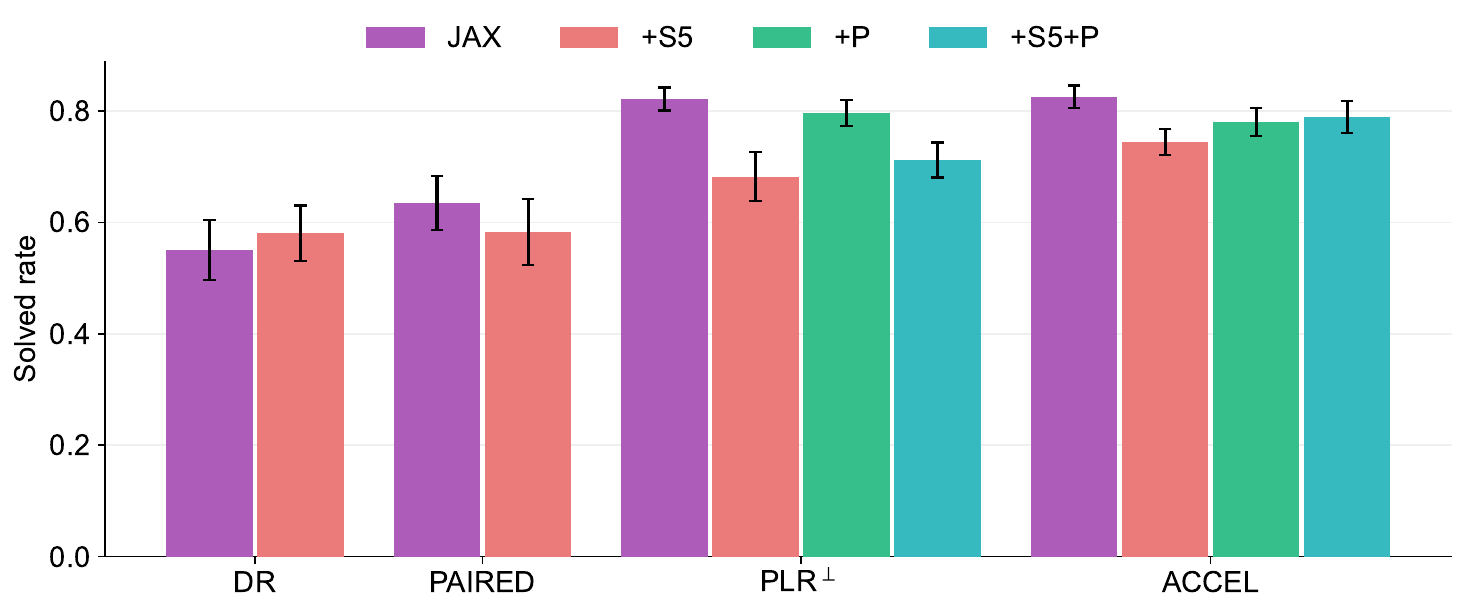}
\caption{Zero-shot transfer solved rate of agents trained via each \texttt{minimax} baseline method averaged across all OOD test mazes (mean and std of 10 runs). This plot shows the absolute performances corresponding to the relative performances reported in Figure~\ref{fig:relative_solved_rate}.} 
\label{fig:benchmark_abs_solved_rate}
\end{figure}

\section{Choice of Architecture and Hyperparameters}
\label{app:hps}

For all experiments, we use PPO~\citep{schulman2017proximal} with generalized advantage estimation~\citep[GAE;][]{schulman2015high} as the base RL optimization algorithm. Our student and teacher policies use the same architecture as described in Jiang et al, 2021~\citep{jiang2021replay}. We report the best hyperparameter settings found for each method with an LSTM policy in Table~\ref{table:hyperparams} and with an S5 policy in Table~\ref{table:hyperparams_s5}. Unless otherwise specified, the teacher and student hyperparameters are equal. For each method we swept subsets of the following hyperparameter values (where applicable) using $5\times$ runs per inspected setting and selected the configuration with the highest mean solved rate over a set of three validation mazes (\texttt{SixteenRooms}, \texttt{Labyrinth}, and \texttt{StandardMaze}---the same used in prior works~\citep{jiang2021replay,parker2022evolving}) after 30k PPO updates:

\begin{itemize}
  \item learning rate in $\{\text{1e-5}, \text{3e-5},\text{5e-5},\text{1e-4},\text{3e-4}, \text{1e-3}\}$
  \item discount factor $\gamma$ in $\{0.999,0.995,0.9\}$
  \item GAE $\lambda$ discount factor in $\{0.95, 0.98, 0.99\}$
  \item student entropy coefficient in $\{0, 0.0001, 0.001, 0.01\}$
  \item teacher entropy coefficient in $\{0, 0.001, 0.01, 0.05\}$
  \item replay rate in $\{0.5,0.8,0.9\}$
  \item PLR scoring function in $\{\text{MaxMC}, \text{PVL}\}$
  \item PLR prioritization in $\{\text{proportional}, \text{rank}\}$
  \item PLR staleness coefficient in $\{0.3,0.5\}$
  \item PLR temperature in $\{0.1,0.3,0.5\}$
  \item ACCEL number of mutations in $\{5,10,20\}$
\end{itemize}

Here, MaxMC refers to the maximum Monte Carlo regret estimator, and PVL, the positive value loss regret estimator~\citep{jiang2021replay}. 

\paragraph{S5 experiments} For S5 policies, we additionally swept over the number of S5 blocks in $\{1,2,4\}$. We set the number of S5 layers to 2, which results in an approximately equal number of model parameters as the LSTM policy used in prior works. Table~\ref{table:hyperparams_s5} reports the best hyperparameters found in combination with an S5 policy.

\paragraph{Batch-size experiments} For our batch size investigations, we swept over the learning rate in $\{\text{3e-5}, \text{1e-4}, \text{3e-4}, \text{1e-3}\}$ while setting the remaining hyperparameters to the best-performing values for 32 parallel environments.

\begin{table}[h!]
\caption{\small{Hyperparameters used for training each method with an LSTM policy.}}
\label{table:hyperparams}
\begin{center}
\scalebox{0.8}{
\begin{tabular}{lrrrrrr}
\toprule
\textbf{Parameter} & DR & PAIRED & \RPLR{} & \PPLR{} & ACCEL{} & \PACCEL{} \\
\midrule
\emph{PPO} & \\
$\gamma$ & 0.995  & 0.995 & 0.999 & 0.999 & 0.999 & 0.999 \\
$\lambda_{\text{GAE}}$ & 0.98 & 0.98 & 0.98 & 0.95 & 0.98 & 0.98 \\
PPO rollout length & 256  \\
PPO epochs & 5 \\
PPO minibatches per epoch & 1  \\
PPO clip range & 0.2  \\
PPO \# parallel environments & 32 \\
Adam learning rate & 1e-4  & 1e-4 & 3e-4 & 3e-4 & 3e-4  & 1e-4 \\
Adam $\epsilon$ & 1e-5  \\
PPO max gradient norm & 0.5  \\
PPO value clipping & yes  \\
return normalization & no  \\
value loss coefficient & 0.5 \\
student entropy coefficient & 1e-3  & 1e-3 & 0.0 & 0.0 & 0.0 & 1e-3 \\
generator entropy coefficient & – & 0.05 & – & – & – & – \\

\addlinespace[10pt]
\emph{\RPLR{}} & & \\
Replay rate, $p$ &  – &  – & 0.5 & 0.5 & 0.8 & 0.8 \\
Buffer size, $K$ &  – &  – & 4000 & 4000 & 4000 & 4000\\
Scoring function &  – &  – & MaxMC & MaxMC & MaxMC & MaxMC   \\
Prioritization &  –  &  – & rank & rank & rank & rank \\
Temperature, $\beta$ &  –  & – & 0.3 & 0.3 & 0.3 & 0.3\\
Staleness coefficient &  – &  – & 0.3 & 0.5 & 0.5  & 0.5 \\

\addlinespace[10pt]
\emph{ACCEL} & & \\
Mutation subsample size, $q$ &  – & – & – & – & 4 & 4  \\
Mutation selection &  – &  – & – & – & batch & batch \\
\# of mutations &  – &  – & – & – & 20 & 20 \\

\bottomrule 
\end{tabular}}
\end{center}
\end{table}

\begin{table}[h!]
\caption{\small{Hyperparameters used for training each method with an S5 policy.}}
\label{table:hyperparams_s5}
\begin{center}
\scalebox{0.8}{
\begin{tabular}{lrrrrrr}
\toprule
\textbf{Parameter} & DR & PAIRED & \RPLR{} & \PPLR{} & ACCEL & \PACCEL{} \\
\midrule
\emph{PPO} & \\
$\gamma$ & 0.995  & 0.995 & 0.999 & 0.999  & 0.995 & 0.999 \\
$\lambda_{\text{GAE}}$ & 0.98 \\
PPO rollout length & 256  \\
PPO epochs & 5 \\
PPO minibatches per epoch & 1  \\
PPO clip range & 0.2  \\
PPO \# parallel environments & 32  \\
Adam learning rate & 3e-5  & 1e-4 & 1e-4 & 1e-4 & 1e-4 & 3e-5 \\
Adam $\epsilon$ & 1e-5 \\
PPO max gradient norm & 0.5 \\
PPO value clipping & yes   \\
return normalization & no   \\
value loss coefficient & 0.5  \\
student entropy coefficient & 1e-3  & 1e-3 & 1e-3 & 1e-3 & 1e-3 & 0.0 \\
generator entropy coefficient & – & 1e-3 & – & – & – & – \\

\addlinespace[10pt]
\emph{\RPLR{}} & & \\
Replay rate, $p$ &  – &  – & 0.5 & 0.5 & 0.5 & 0.8 \\
Buffer size, $K$ &  – &  – & 4000 & 4000 & 4000 & 4000\\
Scoring function &  – &  – & MaxMC & MaxMC & MaxMC & MaxMC \\
Prioritization &  –  &  – & rank & rank & rank & rank \\
Temperature, $\beta$ &  –  & – & 0.5 & 0.5 & 0.1 & 0.1 \\
Staleness coefficient &  – &  – & 0.5 & 0.5 & 0.3 & 0.5\\

\addlinespace[10pt]
\emph{ACCEL} & & \\
Mutation subsample size, $q$ &  – & – & – & – & 4 & 4  \\
Mutation selection &  – &  – & – & – & batch & batch \\
\# of mutations &  – &  – & – & – & 10 & 20 \\

\addlinespace[10pt]
\emph{S5} & & \\
\# layers & 2  \\
\# blocks, $p$ & 2 \\
LayerNorm position & post \\

\bottomrule 
\end{tabular}}
\end{center}
\end{table}

\end{document}